\documentclass[10pt]{article}

\usepackage{times}
\usepackage{epsfig}
\usepackage{graphicx}
\usepackage{amsmath}
\usepackage{amssymb}
\usepackage{xcolor}
\usepackage[breaklinks=true,bookmarks=false]{hyperref}
\usepackage[margin=1.1in]{geometry} 
\usepackage{caption}
\usepackage{titlesec}
\usepackage{abstract}

\usepackage[breaklinks=true,bookmarks=false]{hyperref}
\emergencystretch=\maxdimen
\begin{document}

\title{\Large Leveraging The Finite States of Emotion Processing to Study Late-Life Mental Health }

\author{
Yuanzhe Huang\textsuperscript{1,*} \thanks{These authors contributed equally to this work.} \\ 
yuh94@pitt.edu \and
Saurab Faruque\textsuperscript{2,*} \\ 
faruque.saurab@medstudent.pitt.edu \and
Minjie Wu\textsuperscript{1,2} \\ 
miw75@pitt.edu\and
Akiko Mizuno\textsuperscript{2} \\
akm82@pitt.edu\and
Eduardo Diniz\textsuperscript{1} \\
edd32@pitt.edu\and
Shaolin Yang\textsuperscript{2} \\
yans@pitt.edu\and
George Dewitt Stetten\textsuperscript{1} \\
stetten@pitt.edu\and
Noah Schweitzer\textsuperscript{1} \\
nes93@pitt.edu\and
Hecheng Jin\textsuperscript{1} \\
hej24@pitt.edu\and
Linghai Wang\textsuperscript{1} \\
l.wang@pitt.edu\and
Howard J. Aizenstein\textsuperscript{1,2} \\
aizen@pitt.edu\and
}
\date{} 

\maketitle

\begin{flushleft}
\textsuperscript{1}Department of Bioengineering, University of Pittsburgh, Pittsburgh, Pennsylvania, United States of America  \\
\textsuperscript{2}Department of Psychiatry, University of Pittsburgh School of Medicine, Pittsburgh, Pennsylvania, United States of America 
\end{flushleft}

\maketitle

\section{Introduction}

 Longitudinal behavioral assessments (questionnaires) and Functional MRI are both cornerstones of mental health research.  An ongoing challenge is how best to interpret these temporal patterns of data, i.e., summary behavioral measures and fMRI activity.  Prevailing models process these time-series data using the General Linear Models.  GLM methods assess how the summary scores cross correlate with lags, or how regional fMRI signal correlate with a perceptual stimulus or with other brain regions. These models are useful for some questions.  However, correlations between output signals provide limited insight into the underlying rules (i.e., the controller) that drive the output signals.  It is differences in the underlying controller (the prescription for how to respond to inputs) that fundamentally defines the system.  Finite State Automata (FSA) \cite{automataBook} implemented with Hidden Markov Models (HMM) provide this controller-focused framework.

HMMs have previously been successfully used for behavioral and fMRI analyses. What we present is potentially novel is a simple and intuitive HMM processing pipeline that highlights FSA theory and is applicable for both behavioral analysis of questionnaire data and fMRI data. The questionnaire data is from a national study characterizing older adult mental health changes in response to the COVID-19 pandemic. The fMRI data is on 35 older adults being treated for late-life depression. The data was acquired while participants performed a faces-shapes emotion-reactivity task. We present a series of analyses illustrating how the HMM analysis highlights the brain changes associated with treatment response.

HMMs offer theoretic promise as they are computationally equivalent to finite state automata, the control processor of a Turing Machine. The Turing machine is an abstract computer, made by circumscribing the finite part of the processing to a FSA, which can read and write to an infinite environment (i.e., imaginary computer tape). Modeling the control of the output signals (questionnaire or fMRI) using the FSA provides a way to circumscribe the finite (i.e., explainable) component of the signal.  In this way, the noisy multichannel output signal can be seen as the output of a language, and the HMM encodes the grammar. The Dynamic programming Viterbi algorithm \cite{ViterbiIEEE} is used to leverage the HMM model. It efficiently identifies the most likely sequence of hidden states.  The vcHMM pipeline leverages this grammar to understand how behavior and neural activity relates to depression.   

\section{cvHMM}

We refer to this method as cvHMM because it models the HMM as a simple controller that outputs a change vector.  This change vector is applied to modify the different output signals.  Prior work has applied different framings of the HMM as a controller.  For instance, fMRI studies have instead used autoregressive components to model the changing signals \cite{Dang2017learningEffective} or used the HMM to model the connectivity \cite{Sana2023ANovel}.  Our approach highlights several aspects of FSA theory.  We first describe the method for analyzing fMRI data and then summarize how this applies for analyzing questionnaire data.  

\subsection{cvHMM for fMRI}

If we assume the brain operates as a finite controller, then we know that the brain’s processing over the course of the fMRI acquisition can be described as a sequence of transitions between a finite number of states.  In our framing we think of each state transition as changing the value of the ROIs according to the pre-defined change vector. The FSM models the change score, so there is not a need for an auto regressive model.  The number of states in the FSM model dictates the granularity of the explanation. The existence of a 7-state universal Turing Machine \cite{Minsky1967} shows how powerful a small number of states can be. Since we want to optimize simplicity of the explanation we use a small number of states in our experiments. 

The HMM method takes as input the pre-processed time-series data from 8 ROIs involved in the emotion circuit. Using k-means clustering we induce 7 primary types of change (see figure 1).  We then estimate the probability of moving between these states.  This initializes an HMM capable of generating realistic fMRI time series. In further analyses the HMM is used to induce the hidden state from the initial time-series, rather than the one guessed by the K-means clustering algorithm. 

\subsection{cvHMM for Questionnaires}

Instead of analyzing a fMRI data from 8 ROIs here we build a model describing monthly questionnaire data people filled out about 4 different factors, depression, anxiety, exercise, and loneliness.  The finite controller in this case models finite biological and psychological factors (i.e., rules) that govern how these interact over time.  Just like with the fMRI signal we model the changes over each time point using the change vector assigned to each state of the HMM.  Also, just like with fMRI, k-means clustering is used to induce the change\textunderscore vectors (the different states).  Similarly, the likelihoods of moving between the states (the transition matrix of the HMM) is estimated using the K-Means defined states.  Since we are modeling only 4 signals, we use an even smaller model.  In this case we use a 5 state HMM.

How good the HMM is at modeling the controller can be evaluated by examining whether the induced hidden states are an improvement over the k-means states.  If the induced states enhance the signal, then they can be seen as effectively representing the true nature, and removing noise.  We do a series of comparisons evaluating how these metrics separate fMRI signal of faces versus shapes, and how this relates to depression treatment response prediction. 

\section{Methods}

We implement this cvHMM on fMRI data by using the Gaussian Hidden Markov Model and the Viterbi Algorithm. By defining the number of states from 5 to 9 based on the seven plus minus two theorem, we classify each timepoint of each scan by using the cvHMM for fMRI algorithms. To deal with the label-switching problem of HMM, we implement a majority vote algorithm that calculate the states corresponding to all time points and calculate according to the label whether each state appears more in the time duration that the participants seeing faces or more in the time duration that the participants seeing shape area. If the state appears more in the time duration of faces, we classify this state as faces and mark the corresponding timepoint as 1. Otherwise, we classify the state as shapes and mark the corresponding timepoint as 0. We take the mean value of all 115 scans at each timepoint and get the graph. We also used the K-Means algorithm to cluster data and used the same majority vote algorithm to solve the label-switching problem. 

To investigate the relationship between the length of treatment and the signal between seeing face and seeing shape, we categorized the data into three subsets that are baseline, which is the patient’s first scan, the data scanned after 1 day of the baseline, and 6 days after the baseline. To deal with the label-switching problem of HMM, we use the same majority vote system. 

\section{Data}
\subsection{fMRI Data}
Participants were older adults who were recruited for a treatment trial of a current depressive disorder.  They were enrolled in an RCT of 2 different anti-depressants (escitalopram versus levomilnacopran).  The sample and protocol have recently been described in our report demonstrating prediction of treatment response. The task and study protocol has been described in our prior reports \cite{aizenstein2011} \cite{khalafAltered}. Data was acquired at 4 different days over the course of a 12-week trial.  Data is from 34 participants, with up to 4 different time points per participant (i.e., 115 scans session).  

The task is the faces-shapes affective-reactivity task on which we have previously published.  Participants are shown a series of affective faces (angry or fearful) 2 on top and one below.  They are asked to press a key according to which of the 2 faces above matches the one face below. Blocks of faces stimuli are interleaved with blocks of trials in which participants match simple abstract shapes. The task is known to highlight circuits for emotion reactivity and emotion regulation.  Our ROI-based analytic approach focuses on 8 ROIs important in emotion reactivity and emotion regulation: bilateral regions in the anterior cingulate, dorsolateral prefrontal cortex, Anterior Insula, and Amygdala. 

We have previously published standard GLM analysis with this same task and the same treatment-trial design \cite{aizenstein2011} \cite{karimht2018}.  We have also recently described how the 7T fMRI data at rest is associated with treatment response\cite{acute2}. 

\subsection{Questionnaire data and sample}
The COVID-19 Coping Study is a national, longitudinal cohort study of 6,938 US adults aged $>=$55 during the first year of the COVID-19 pandemic in order to collect data on the psychosocial and behavioral impact of the pandemic on older adults. Of the 6,938 subjects included in COVID-19 Coping Study, 823 subjects had complete data for our four selected variables of interest at all collected timepoints. 

\subsubsection{Mental health measures}
The COVID-19 Coping Study collected mental health data using validated questionnaires, including the Center for Epidemiological Studies Depression (CES-D) Scale, three-item UCLA Loneliness Scale, and Beck Anxiety Inventory (BDI). 
Depression was measured using the eight-item Center for Epidemiological Studies Depression (CES-D) Scale. The prompt asked if feelings were experienced “much of the time” in the past week, with “yes” and “no” scored as 1 and 0 respectively. The eight items were “you felt depressed,” “you felt that everything you did was an effort,” “your sleep was restless,” “you were happy,” “you felt lonely,” “you enjoyed life,” “you felt sad,” and “you could not get going.” The items “you were happy” and “you enjoyed life” were reverse-coded. The final measure was the sum of the eight items. Subjects were further categorized as “depressed” if their CES-D score was $>$ 3 and “not depressed if $<$ 3. 

Loneliness was measured using the three-item University of California, Los Angeles (UCLA) Loneliness Scale. The items asked how often the subject felt “left out,” “lacked companionship” and “isolated” during the past week, with the response options “hardly ever,” “some of the time,” and “often” scored 1 to 3 points respectively. The final measure was the sum of the three individual items. 

Anxiety was measured using items from the Beck Anxiety Inventory. The measure was collected by prompting: “After each statement, please indicate how often you felt that way during the past week,” with the response options “never,” “hardly ever,” “some of the time,” and “most of the time” scored 1 to 4 points respectively. The 5 statements were: “I had a fear of the worst happening,” “I was nervous,” “I felt my hands trembling,” “I had a fear of dying,” and “I felt faint.” The final measure of anxiety was the average of the 5 individual items. 

Exercise was measured using the item "during the past week, how much moderate-to-vigorous exercise did you do (e.g., exercise for leisure, transportation, or housework that gets your heart rate elevated and makes you breathe faster)?” with “none” scored as 0, “$<$30 min” scored as 1, “30 min to $<$1 hour” scored as 2, “1 hour to $<$1.5 hours” scored as 3, “1.5 hours to $<$2 hours” scored as 4, “2 hours to $<$2.5 hours” scored as 5, and “2.5+ hours” scored as 6. Exercise was collected every other month, so uncollected months were interpolated using mean imputation using adjacent timepoints.  

Baseline timepoint data was not included in the study because inter-subject variance in the interval between baseline and timepoint 1 while the remainder of the data had a consistent collection interval of 1 month between timepoints 1-12. 
Z-score transformation was applied to each variable across all subjects and timepoints of Mental health measure standardization.

\section{Results}
\subsection{fMRI Data}

Our result that implement cvHMM on fMRI data shows that the Viterbi algorithm can clean up the noise. Figure 1 shows the result of mean value of fMRI raw data of bilateral amygdala (Green Line), and the mean value of Viterbi Algorithm and K-Means. The mean value is calculated by using the majority vote algorithm that we discussed in previous part. The Red line in the figure 1 is the label that if red line is 1 the participants saw faces, otherwise the participants saw shapes. In table 1 we provide the effect size of Viterbi Algorithm, K-Means, and Bilateral Amygdala by using the data scans of Baseline, day 1 after baseline, and day 6 after baseline. From table 1 we observe that the effect size of Bilateral Amygdala is showing similar pattern of Viterbi that Day 1 has the highest effect size and Baseline has the lowest effect size. In table 2, we also provide a table of participants' age, sex and their depression score at their baseline scan and final point scan. 

\begin{figure}
    \centering
    \includegraphics[width=16cm]{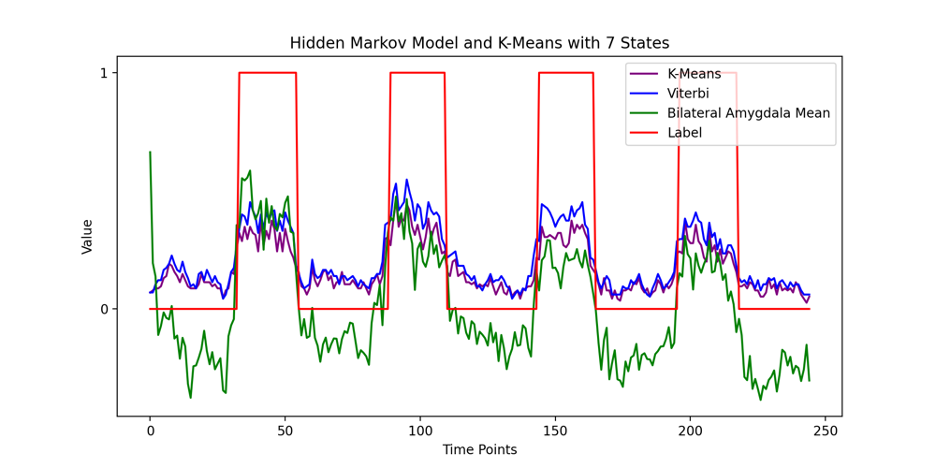}
    \caption{The mean of bilateral amygdala, Viterbi, and K-Means that use all 115 scans data }
    \label{fig:enter-label}
\end{figure}

\begin{table}[h!]
\centering
\caption{Effect size of data of baseline, Day 1, and Day 6 of Viterbi, K-Means, and Bilateral Amygdala }
\label{tab:scanner_specs}
\begin{tabular}{llcc}
\hline
Data & Viterbi & K-Means & Bilateral Amygdala \\
\hline
Baseline & 1.6502 & 1.6383 & 1.4905 \\
Day 1 & 1.7064 & 1.3994 & 1.5569 \\
Day 6 & 1.6841 & 1.5735 & 1.5446 \\
\hline
\end{tabular}
\end{table}

\begin{table}[h!]
\centering
\caption{Table of Participants' Age, Sex, and Depression score at Baseline and Final Point}
\label{tab:scanner_specs}
\begin{tabular}{ c c c c }
\hline
Sex & Age & Baseline & Final \\
\hline
Male & 79 & 18 & 12 \\
Male & 60 & 34 & 25 \\
Male & 69 & 21 & 2 \\
Female & 67 & 17 & 1 \\ 
Male & 82 & 20 & 20 \\
Female & 74 & 30 & 30 \\ 
Female & 75 & 34 & 6 \\
Male & 69 & 16 & 10 \\
Female & 62 & 25 & 0 \\ 
Female & 67 & 20 & 2 \\
Male & 63 & 27 & 5 \\
Female & 63 & 29 & 0 \\
Female & 74 & 16 & 6 \\
Male & 70 & 18 & 4 \\
Male & 63 & 25 & 5 \\
Female & 61 & 18 & 0 \\
Female & 68 & 18 & 6 \\
Female & 62 & 21 & 11 \\
Male & 63 & 27 & 11 \\
Male & 66 & 14 & 0 \\
Male & 60 & 17 & 14 \\
Male & 65 & 15 & 9 \\
Female & 72 & 15 & 7 \\
Female & 71 & 20 & 9 \\
\hline
\end{tabular}
\end{table}

\subsection{Questionnaire Data}
\subsubsection{Questionnaire change-vectors}
Change-vectors represent the magnitude and direction of change in one variable from each timepoint t to t+1. For each variable, z-score standardized values were used to calculate change-vectors. 12 original timepoints yield 11 change-vector timepoints, and thus 11 change-vectors for each of the four variables of interest (depression, loneliness, anxiety, and exercise). 

\subsubsection{Change-vectors and change-vector-patterns}
If each variable is seen as a single dimension reflective of a subject’s underlying psychological state, each change-vector for a certain variable can be seen as a 1-dimensional feature of an underlying multidimensional change-vector. Thus, the combination of each of the levels of the four 1-dimensional change-vectors at each timepoint could be combined to represent a single change-vector in a 4-dimensional space. The 4-dimensional change-vector is more aptly described named a ‘change-vector-pattern' as it was derived from the unique combinations of the four 1-dimensional change-vectors for each subject. 

\subsubsection{K-means induced change-states}

\begin{figure}
    \centering
    \includegraphics[width=14cm]{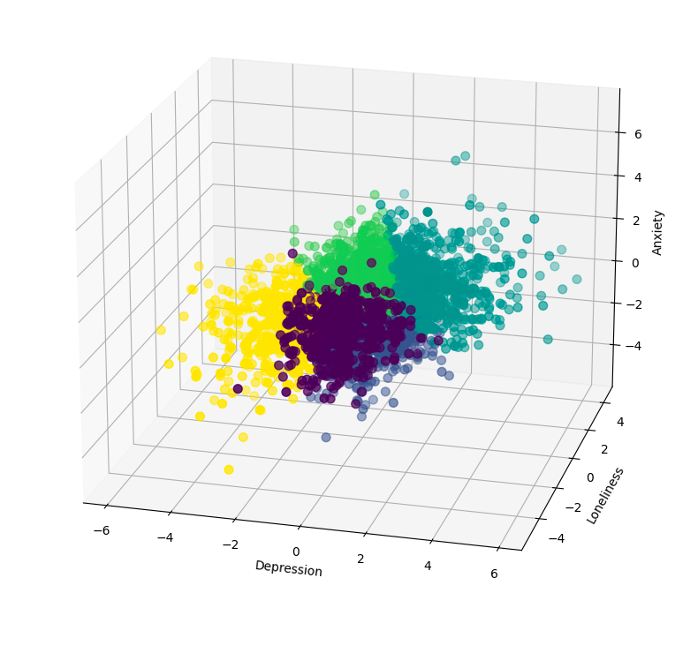}
    \caption{3-dimensional representation of k-means induced change-states, excluding exercise variable }
    \label{fig:enter-label}
\end{figure}

\begin{figure}
    \centering
    \includegraphics[width=16cm]{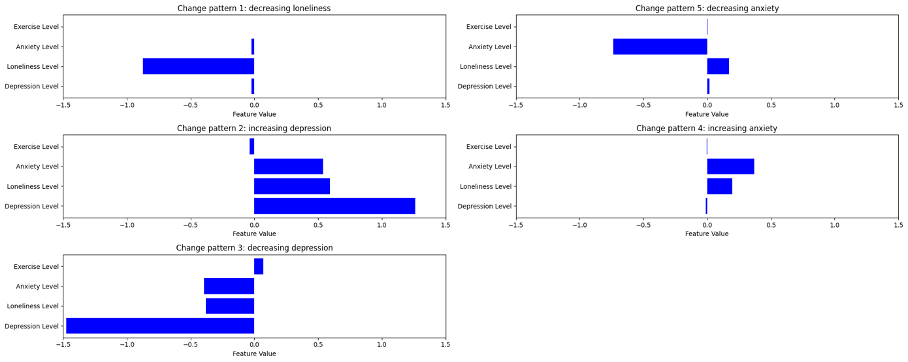}
    \caption{Change-states defined by salient factor (i.e., its prominent contributory change-vector dimension) }
    \label{fig:enter-label}
\end{figure}

The k-means clustering algorithm used the change-vector-patterns for each subject at every timepoint to generate five clusters in a 4-dimensional space. Figure 2 presents a visualization of a 3-dimensional k-means clustering excluding exercise. K-means required 23 iterations for convergence with four variables. Each of the five 4-dimensional change-vector-pattern clusters generated by k-means have a representative centroid, which will be termed change-states, to which each subject can be classified at every timepoint. Each change-state has a distinctive dimension—termed salient factor—reflecting the prominent contributory variable. The five generated change-states were characterized by the following salient factors: 1) decreasing loneliness, 2) increased depression, 3) decreased depression, 4) increasing anxiety, and 5) decreasing anxiety (Figure 3). Assigning each subject a change-state at each timepoint creates a change-state sequence for that subject. 

The change-state sequences for each subject can then be investigated individually to characterize the frequency of their transitions between two temporally adjacent change-states (i.e., the number of times a subject transitioned between any two change-states at any adjacent sequential timepoint pairs). When done for all subjects at every timepoint, we can derive the likelihood of transitioning from any one change-state to any other change-state from the frequency of those transitions. 
\begin{figure}
    \centering
    \includegraphics[width=16cm]{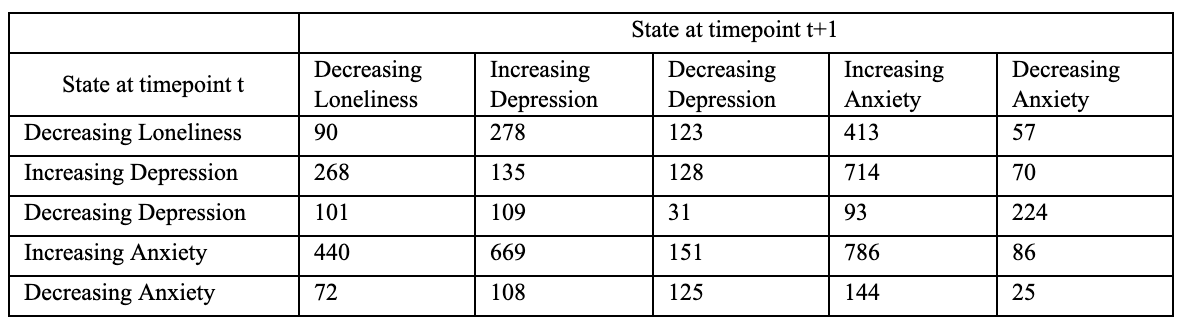}
    \caption{Frequencies of change-state transitions for female subjects in k-means induced sequences. }
    \label{fig:enter-label}
\end{figure}

\subsubsection{Viterbi generated change-state transitions}
Change-state transition likelihoods, change-vectors, and parameters induced by the k-means algorithm (i.e., change-states) can then be inputted to the Viterbi algorithm. Viterbi is a dynamic programming algorithm that conducts an exponential search to estimate the Viterbi path—the most likely sequence of hidden states—to output a sequence of observed events. The Viterbi generated HMM change-state sequences and the k-means induced change-state sequences can be compared by examining the frequencies of their change-state transitions. The frequencies of change-state transitions for any group or subgroup in the sample can be represented in a 5x5 matrix that captures every combination of change-state transitions (e.g., Tabel in Figure 4). \\

\subsubsection{Comparison of Viterbi-generated and k-means induced change-state transitions}
The change-state transition matrices for different groups can then be compared using chi-squared test of independence, with the null hypothesis being that there is no association between the group and the change-state transition patterns. Cramér's V can be calculated to measure the strength of the association between the groups and the change-state transition patterns, with a higher value indicating a stronger association. We examined the effect of Viterbi induction on the change-state transition patterns of gender and depression status. Tables in the Figure 5-8 show the chi-squared residuals when comparing gender (Table 5 and 6) and depression status (Table 7 and 8). Black filling represents change-state transitions that were not significantly different between groups (chi-squared residual $<$ 2). All other colored change-state transitions were statistically significant, with darker shading indicating a larger residual (green if positive and red if negative). 

\begin{figure}
    \centering
    \includegraphics[width=8cm]{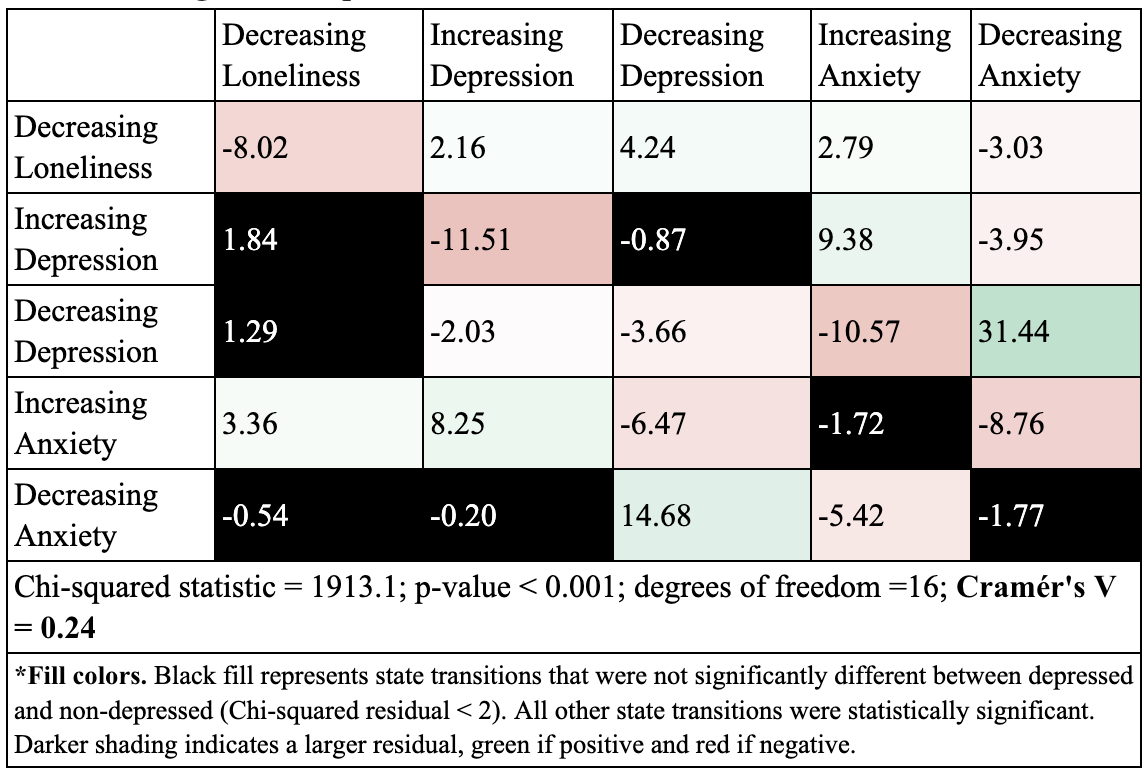}
    \caption{Chi-squared residuals for male and female state transition frequencies in K-means induced change-state sequences. }
    \label{fig:enter-label}
\end{figure}
\begin{figure}
    \centering
    \includegraphics[width=8cm]{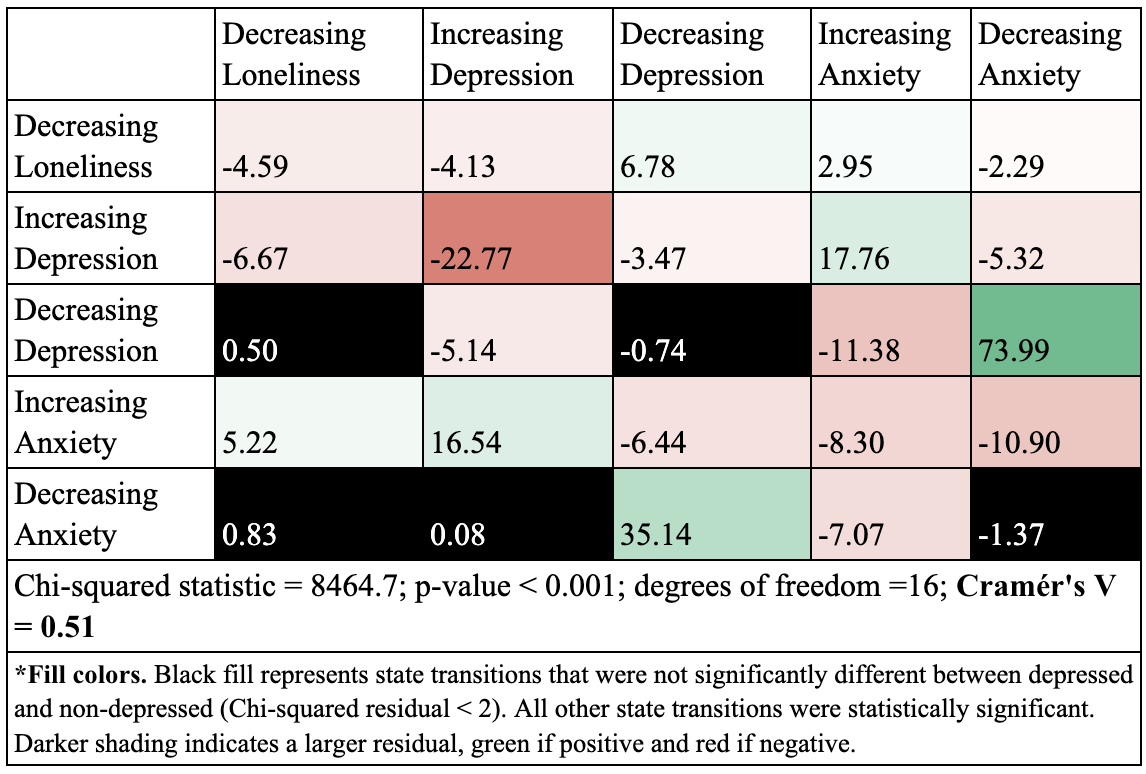}
    \caption{Chi-squared residuals for male and female subjects state transition frequencies in Viterbi generated change-state sequences. }
    \label{fig:enter-label}
\end{figure}
\begin{figure}
    \centering
    \includegraphics[width=8cm]{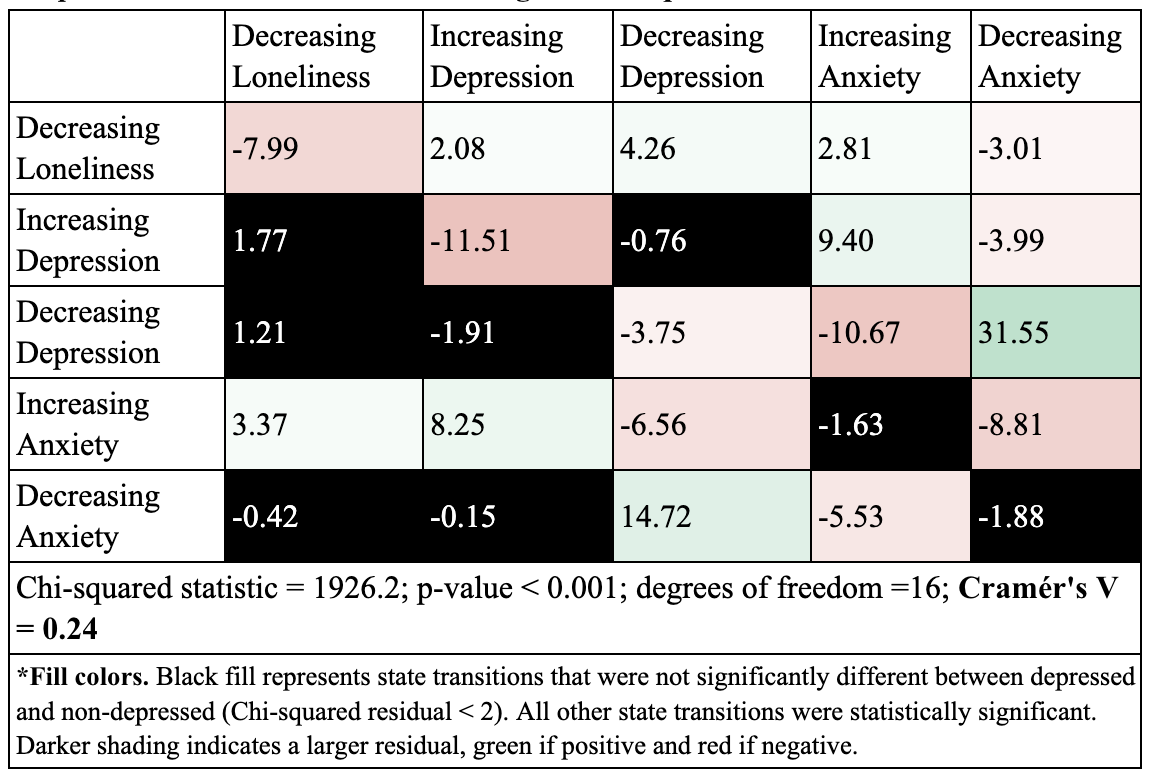}
    \caption{Chi-squared residuals for depressed and non-depressed subjects state transition frequencies in K-means induced change-state sequence. }
    \label{fig:enter-label}
\end{figure}
\begin{figure}
    \centering
    \includegraphics[width=8cm]{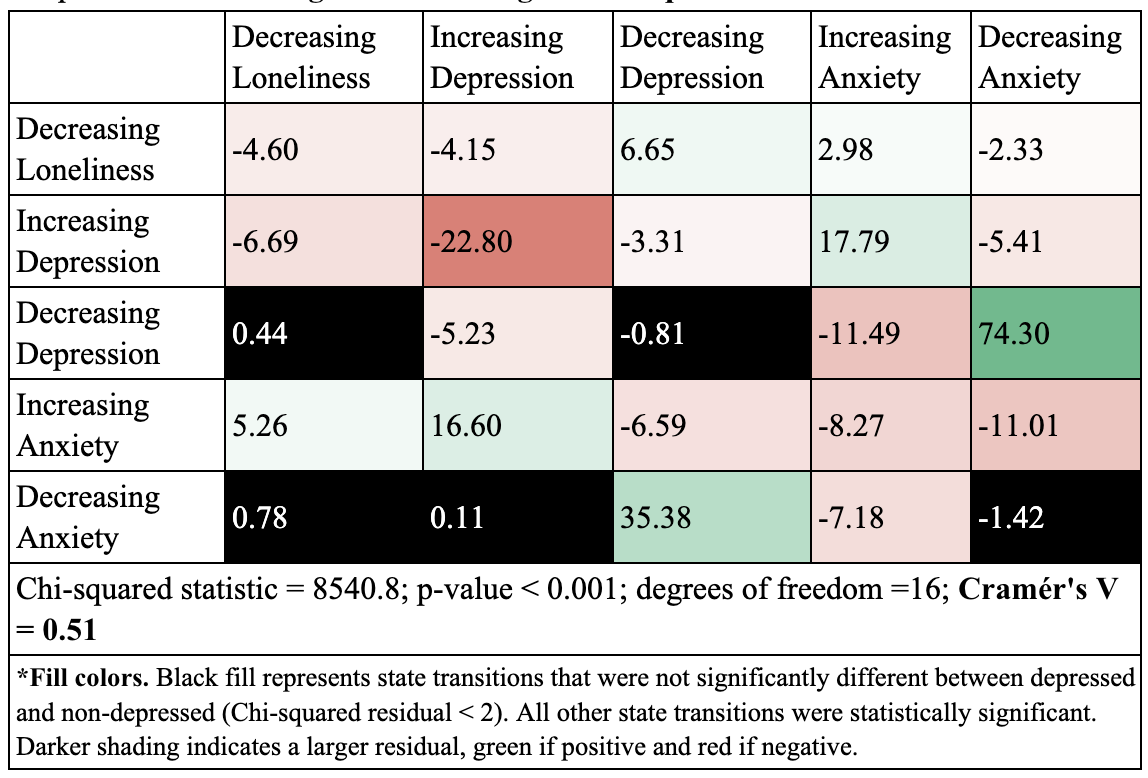}
    \caption{Chi-squared residuals for depressed and non-depressed subjects state transition frequencies in Viterbi generated change-state sequences. }
    \label{fig:enter-label}
\end{figure}

For both depression status and gender, the Viterbi generated change-state sequence demonstrates a stronger association, as indicated by a higher Cramér's V, compared to the k-means induced change-state sequence. This suggests that the sequences decoded by the Viterbi algorithm from the Hidden Markov Model capture a more pronounced relationship between these variables and change-state transition frequencies, resulting in a clearer distinction between the signal (the association) and the noise (random variability). In other words, the variations in transition frequencies related to depression status and gender are more distinct and less influenced by unexplained or random variability in the Viterbi generated sequences. 

\section{Discussion}

In this paper we have described an intuitive pipeline for using the HMM framework for modeling behavior and fMRI.  The model highlights the relationship between the HMM and the Turing Machine (TM).  The TM separates any processor (finite or infinite) into 2 separate components: a finite, potentially explainable component, and an infinite read/write tape.  This formulation puts the focus on the natural patterns in the processing that are constrained by the finite aspects of the controller.

The finite controller of the TM is sometimes referred to as Finite State Automate (FSA).  The limited number of states force some repeatable pattern in the data.  These patterns, can also be described as a language.  One of the key finding in automata and language theory \cite{automataBook} is the equivalence of FSAs with regular expression grammars.  Thus, we can think of this approach as highlighting the true signal by putting it into meaningful word types (i.e., in this case states).

The FSA is computationally equivalent to an HMM.  Therefore, in following the TM analogy, we restrict components of the fMRI or behavioral signal to a type of HMM that serves as a controller. The change vector HMM (cvHMM) framework highlights this formulation.  The HMM is modeled as a simple controller, with a data-driven control signal.  Modeling the signal as a simple change vector is a reasonable first pass.  A more general data-driven change function, such as a Multilayer Perceptron network that reads and writes to the context would likely provide a more nuanced fit. Moreover, thinking of the change vector as a general function manipulating a representation space is even more appealing.  In this way, the representation space is what moves over time, and the states are data-driven functions that read-write to the time series of representation spaces.  This whole process can be done iteratively to find a cvHMM that most enhances the signal on some pretext condition.

The cvHMM framework helps characterize a principled higher level knowledge representation (structured representation space) for studying causal (often temporal) data.  The FSA model restricts the possibilities to those that come from the finite explainable part.  It leverages the notion that machines are languages.  This frames questions about behavior or neural processing, as also being questions about the underlying languages.  The vcHMM framework is one simple HMM expression of that grammar.  The model provides an alternative to correlation-based approaches of fMRI.  It also provides a generative model.  This can be leveraged in deep learning to further enhance the signal.  This can provide a nuanced understanding of the dynamics underlying mental health, particularly in the context of late-life depression.

For fMRI data analysis, we proposed that the brain functions as a finite controller, with transitions between a finite number of states dictating its processing during fMRI acquisition. By employing k-means clustering, we induced primary types of change and estimated the probabilities of transitioning between these states, thereby initializing an HMM capable of generating realistic fMRI time series. This enabled us to highlight the brain changes associated with treatment response in older adults with late-life depression. 

In the case of questionnaire data, we constructed a model that describes monthly data on depression, anxiety, exercise, and loneliness. Here, the finite controller represents the biological and psychological factors that govern the interactions of these variables over time. Using a similar approach as with the fMRI data, we employed k-means clustering to induce change vectors and estimate transition probabilities between states. 

Our results demonstrated the effectiveness of the HMM in modeling the controller, as evaluated by comparing the induced hidden states with the k-means states. We found that the HMM could enhance the signal by accurately representing the true nature of the data and effectively removing noise. This was further supported by comparisons of change-state transition frequencies, which showed a clearer distinction between signal and noise in the Viterbi-generated sequences compared to the k-means induced sequences. 

Moreover, our approach has several advantages over traditional methods. Firstly, it provides a more granular view of the data, allowing for the identification of subtle changes in mental health states. Secondly, it offers a robust framework for analyzing both fMRI and questionnaire data, which can be applied to other populations and mental health conditions. Thirdly, the cvHMM approach is flexible and can be integrated with other computational methods, providing a comprehensive tool for mental health research. 

However, there are also some limitations to our study. The accuracy of the model depends on the quality of the input data and the appropriateness of the clustering algorithm. Additionally, the interpretation of the results requires a deep understanding of both HMM and FSA theory, which may be challenging for some researchers. 

In our approach, we have emphasized the importance of simplicity in the explanation of mental health data. This is inspired by Marvin Minsky's concept of a 7-state universal Turing machine, which demonstrates how a small number of states can be powerful enough to represent complex systems.  By using a small number of states in our experiments, we aim to optimize the simplicity of the explanation, making it more manageable and interpretable for researchers and clinicians. 

In conclusion, our study highlights the potential of combining HMM with FSA theory for a deeper understanding of the dynamics of mental health in older adults. The cvHMM approach provides a robust and flexible framework for analyzing both fMRI and questionnaire data, offering new insights into the brain's processing and the factors influencing mental health. Future research should explore the application of this approach to other populations and mental health conditions, as well as its integration with other computational methods. This could lead to the development of more effective treatments and interventions for mental health disorders.


\bibliographystyle{unsrt}
\bibliography{bibliography.bib}

\end{document}